\documentclass[letterpaper, 10 pt, conference]{ieeeconf}  
\IEEEoverridecommandlockouts
\usepackage{graphicx}
\usepackage{amssymb,amsmath,physics}
\usepackage{array,booktabs,arydshln}
\usepackage{algorithm}
\usepackage{algorithmic}
\usepackage[utf8]{inputenc}
\usepackage[english]{babel}
\usepackage{xcolor}
\definecolor{green}{RGB}{3,112,15}
\definecolor{yellow}{RGB}{255,140,0}
\usepackage{hyperref}
\usepackage{cite}
\usepackage{bm}
\usepackage{sidecap}
\usepackage{subcaption}
\usepackage{subcaption}
\usepackage{cleveref}

\newcommand{\saad}[1]{#1}

\newcommand{\acronym}{LfH-CP}

\DeclareMathOperator*{\argmax}{argmax}
\DeclareMathOperator*{\argmin}{argmin}

\title{\bf
Learning from Hallucinating Critical Points for \\Navigation in Dynamic Environments
}

\author{Saad Abdul Ghani$^{1}$, Kameron Lee$^{2}$, and Xuesu Xiao$^{1}$
\thanks{
$^{1}$George Mason University {\tt\small \{sghani2, xiao\}@gmu.edu}
}
\thanks{$^{2}$Thomas Jefferson High School for Science and Technology intern affiliated with the RobotiXX Lab, George Mason University. {\tt\small 2027klee@tjhsst.edu}.}
}

\begin{document}
\bstctlcite{IEEEexample:BSTcontrol}

\maketitle
\begin{abstract}

Generating large and diverse obstacle datasets to learn motion planning in environments with dynamic obstacles is challenging due to the vast space of possible obstacle trajectories. Inspired by hallucination-based data synthesis approaches, we propose Learning from Hallucinating Critical Points (\acronym), a self-supervised framework for creating rich dynamic obstacle datasets based on existing optimal motion plans without requiring expensive expert demonstrations or trial-and-error exploration. \acronym~factorizes hallucination into two stages: first identifying when and where obstacles must appear in order to result in a near-optimal motion plan, i.e., the critical points, and then procedurally generating diverse trajectories that pass through these points while avoiding collisions. This factorization avoids generative failures such as mode collapse and ensures coverage of diverse dynamic behaviors. We further introduce a diversity metric to quantify dataset richness and show that \acronym~produces substantially more varied training data than existing \saad{baseline}. Experiments in simulation demonstrate that planners trained on a \acronym~generated dataset achieves higher success rates compared to a prior hallucination method.
\end{abstract}

\section{Introduction}
\label{sec::intro}

Dynamic obstacles present a fundamental challenge for autonomous mobile robots, as their trajectories reside in a vast and high-dimensional space. Describing obstacle trajectories with velocity, acceleration, and higher order derivatives adds dimensions to the space, making it difficult for planners to anticipate and respond to complex motion patterns in real time.

Learning-based models have recently demonstrated success in navigating such environments by leveraging collected data~\cite{xiao2022motion}. Two dominant paradigms—Imitation Learning (IL) and Reinforcement Learning (RL)—provide structured ways to gather and learn from experience. However, both paradigms face critical limitations: IL requires large numbers of expert demonstrations, while RL demands extensive trial-and-error exploration. Moreover, robust planners depend on training datasets that are sufficiently diverse to capture the variability of dynamic obstacles. As a result, learning-based planning has not performed to the same extent as in vision and language domains, where internet-scale labeled datasets are readily available.

\begin{figure}
  \centering
  \includegraphics[width=0.8\columnwidth]{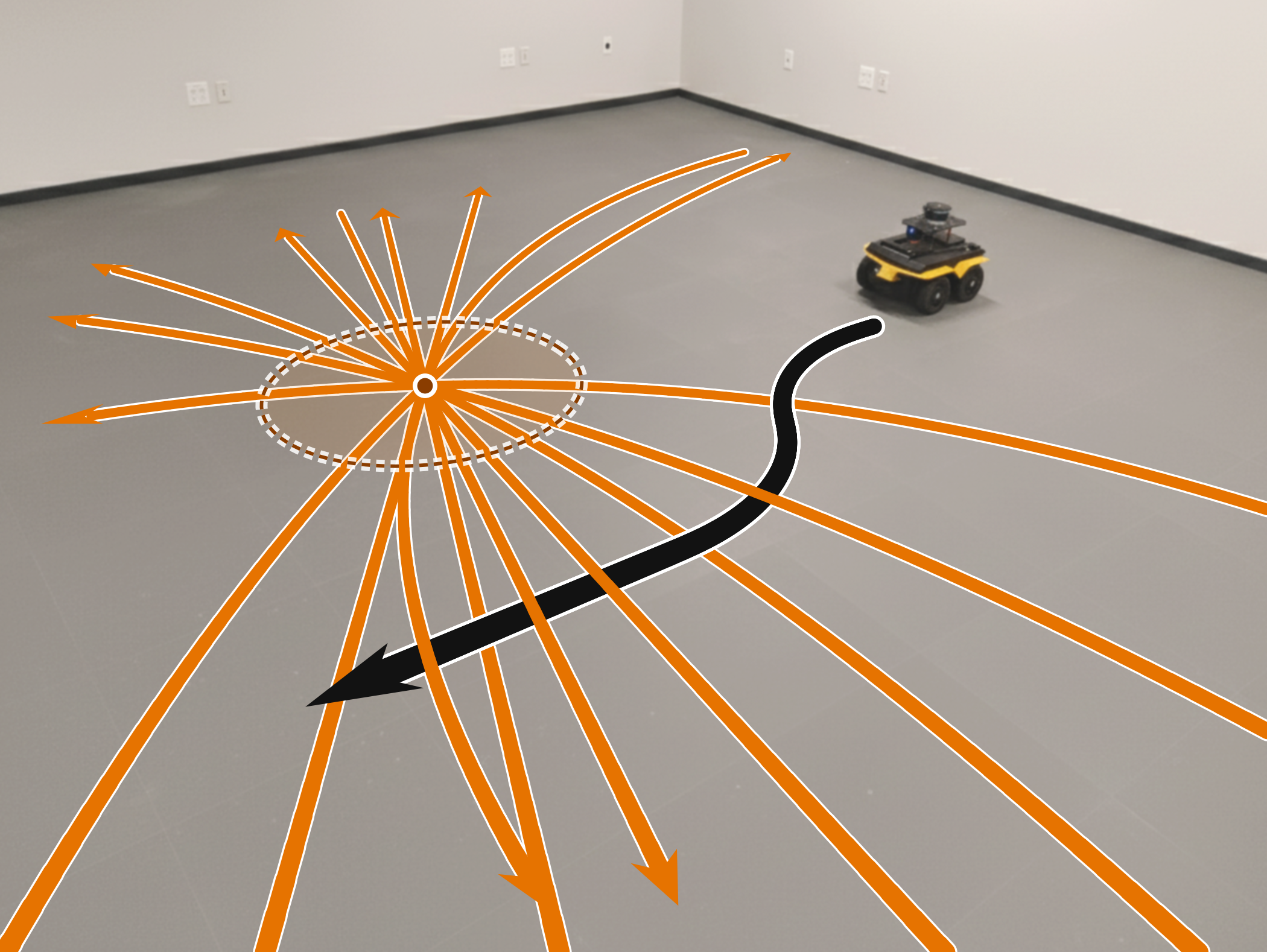}
  \caption{A single hallucinated critical point (dotted circle) — a location and time at which an
obstacle must appear for the robot's plan (black) to be near-optimal. Any obstacle trajectory
passing through that point (orange arrows) leaves the same plan valid, so one critical point
yields many distinct training scenarios.}
  \label{fig::dynaInAction}
  \vspace{-10pt}
\end{figure}

The Learning from Hallucination (LfH) paradigm~\cite{xiao2021toward} addresses this by synthesizing training data from prior navigation experiences. LfH takes past motion plans from simpler, even fully open environments and generates more constrained and complex scenarios, where those prior plans would be optimal. This process, known as ``hallucination'', enables safe and inexpensive creation of large datasets without relying on expensive expert demonstrations or extensive trial-and-error exploration. However, prior methods often produce datasets with limited diversity, constraining their applicability and scalability, especially when facing the vast and high-dimensional space of dynamic obstacles~\cite{wang2021agile, xiao2021agile}.

\begin{figure*}[t]
  \centering
  \begin{subfigure}{1.0\columnwidth}
    \includegraphics[width=\linewidth]{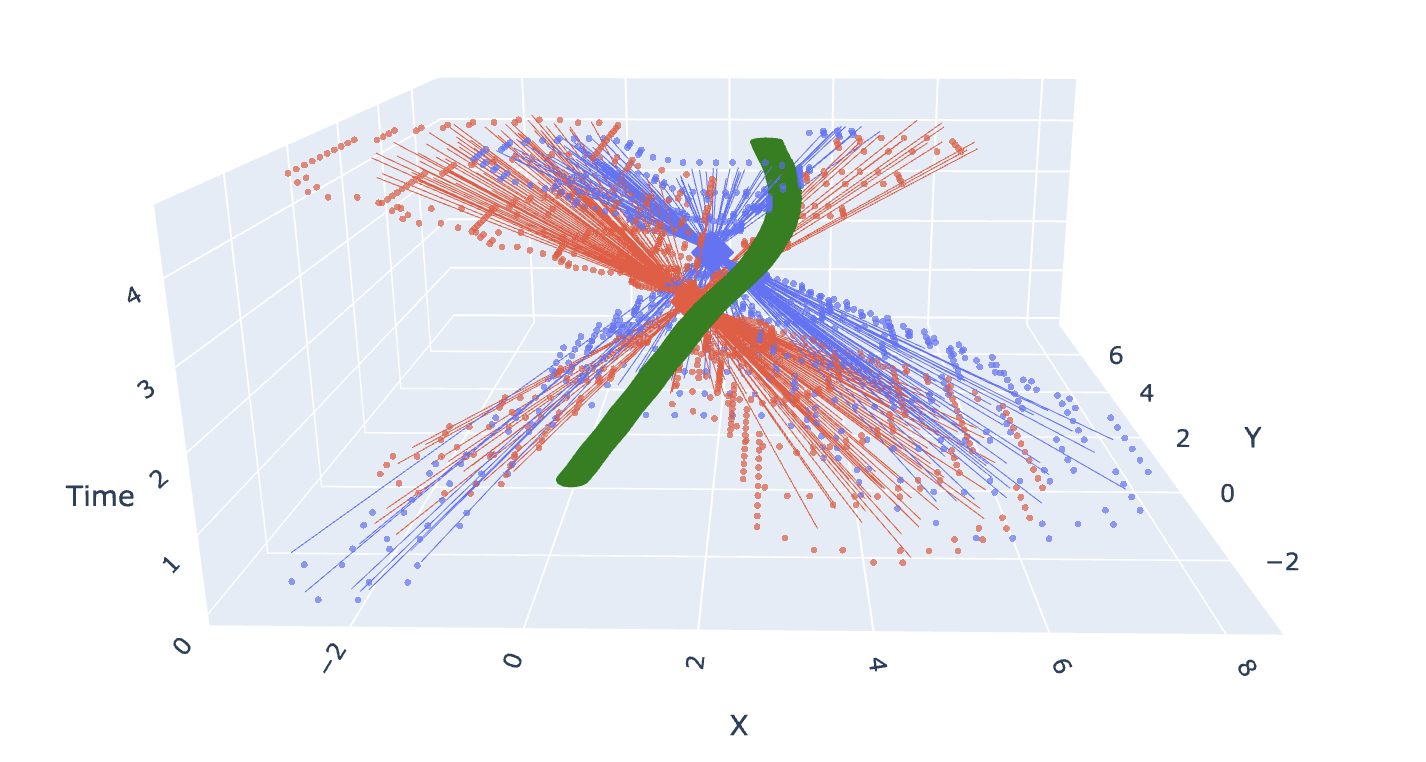}
    \caption{}
    \label{fig:A}
  \end{subfigure}
  \hfill
  \begin{subfigure}{1.0\columnwidth}
    \includegraphics[width=\linewidth]{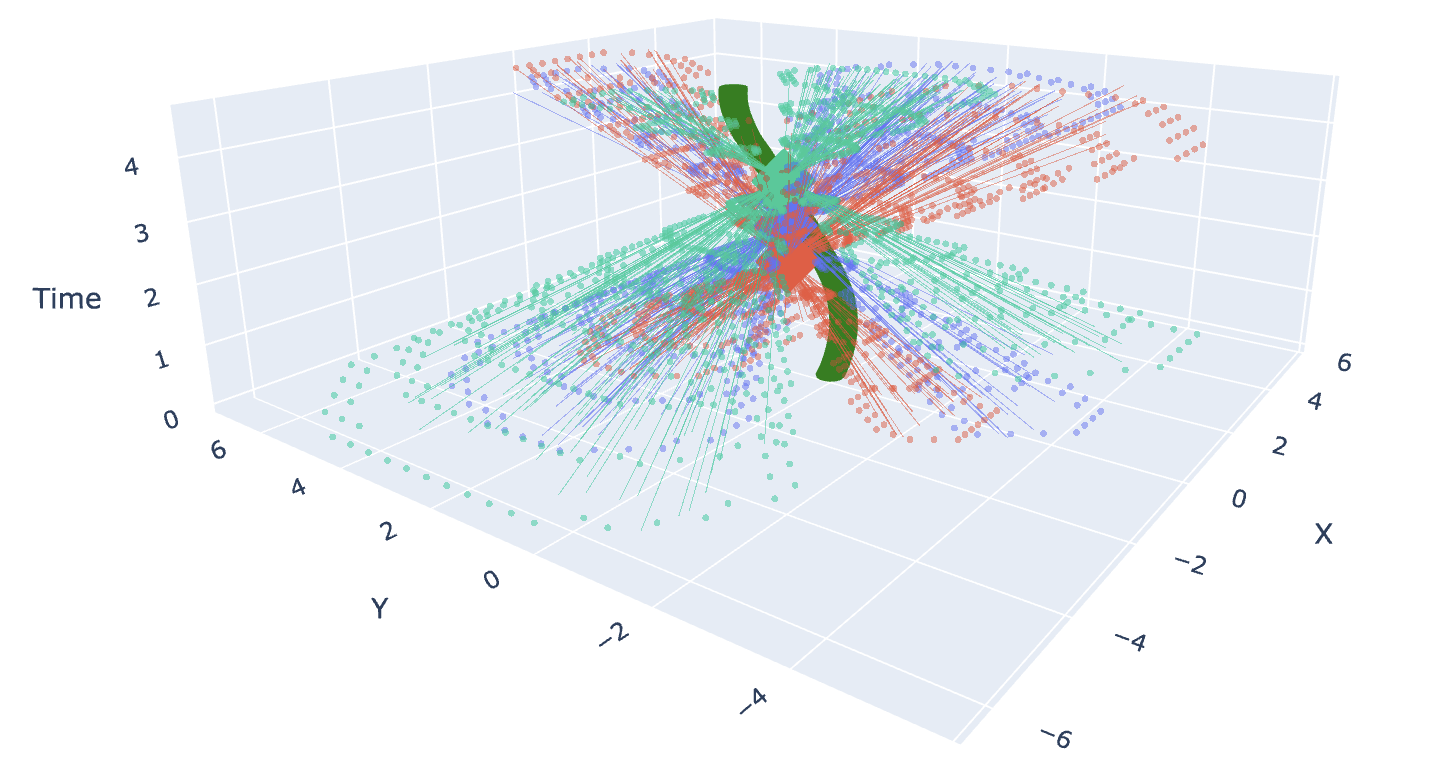}
    \caption{}
    \label{fig:B}
  \end{subfigure}
  \caption{
  \small
  100 obstacle trajectories are generated from each critical point. The robot trajectory (green) remains collision-free and near-optimal, regardless of the variations and combinations of obstacle trajectories. That is, the same robot trajectory is near-optimal and collision-free for any single combination of these $100^N$ possible obstacle trajectory sets. Fig. \ref{fig:A} shows a total of 200 obstacle trajectories generated from $N=2$ critical points, while Fig. \ref{fig:B} shows a total of 300 obstacle trajectories generated from $N=3$ critical points.}
  \label{fig::generated_obstacle_trajectories}
\end{figure*}

In this work, we introduce Learning from Hallucinating Critical Points (\acronym), a self-supervised framework for generating rich datasets of dynamic obstacle trajectories. \acronym~ factorizes hallucination into two stages: first identifying the \textit{critical points} of obstacle trajectories—specific times and locations obstacles must appear in order to result in a near-optimal motion plan—and then procedurally generating diverse trajectories with varying orders of complexity that pass through these points while avoiding collisions (Fig.~\ref{fig::generated_obstacle_trajectories}). These generated obstacle trajectories and original motion plans can then be used to train motion planners in a supervised manner.
To quantify the richness of the generated datasets, we introduce a diversity metric and show that \acronym~produces substantially more varied training data than existing methods. We validate these results in simulated experiments on a ground robot, demonstrating that planners trained on \acronym~datasets achieve superior navigation performance compared to a prior hallucination-based \saad{baseline}~\cite{ghani2025dyna}.
In summary, this work makes the following contributions:
\begin{itemize}
\item Propose \acronym, a self-supervised framework for generating large and diverse datasets of dynamic obstacles.
\item Introduce a diversity metric to quantify dataset richness.
\item Demonstrate effectiveness of \acronym~through simulated robot experiments.
\end{itemize}

\section{Related Work}
\label{sec::related}

This section reviews learning based techniques used in autonomous mobile robots for navigating in dynamic environments. It also reviews the recent LfH paradigm.

\subsection{Machine Learning for Dynamic Obstacle Avoidance}
Machine learning has been leveraged in mobile robot navigation in various ways~\cite{xiao2022motion, xiao2026autonomous}, such as either integrating learning with classical methods~\cite{semnani2020multi, xu2021applr} or using IL~\cite{pfeiffer2017perception, xiao2022learning} or RL~\cite{xu2023benchmarking, xu2021machine} to develop end-to-end planners~\cite{zeng2019navigation, wullt2023neural}. 
Going beyond simple obstacle avoidance~\cite{xiao2022autonomous, xiao2024autonomous, liu2021lifelong}, learning is heavily used in social robot navigation where classical methods fail to calculate trajectories for human-populated spaces~\cite{xiao2022learning,  nguyen2023toward, francis2025principles}. 
Another use case is off-road navigation, in which learning can help to reason about the unstructured terrain underneath the robot~\cite{karnan2022vi, datar2024terrain, datar2024learning, pokhrel2024cahsor}. 

However, despite their success, most learning methods require either high-quality (IL)  or extensive (RL) training data, such as those derived from human demonstrations or from trial-and-error exploration respectively, both of which are very difficult to acquire among dense and dynamic obstacles. \acronym\ is a self-supervised learning approach that automatically generates diverse training data, addressing the conundrum of needing to know what good navigation behavior is, without prior knowledge of how to achieve it.

\subsection{Learning from Hallucination (LfH)}
LfH~\cite{xiao2021toward, xiao2021agile, wang2021agile, park2023learning, ghani2025dyna} has been proposed to alleviate the difficulty of acquiring high-quality or extensive training data using completely safe exploration in open spaces or existing successful navigation experiences. Based on existing motion plans, the idea of hallucination is to generate obstacle configurations in which the existing plans would be optimal. Hallucination allows robots to reflect on past success and produces other training scenarios where such successful navigation can be repeated. Researchers have designed hallucination techniques to project the \emph{most constrained}~\cite{xiao2021toward}, a \emph{minimal}~\cite{xiao2021agile}, or a \emph{learned}~\cite{wang2021agile, ghani2025dyna} obstacle configuration onto the robot perception. Hallucination has also been used to enable multi-robot navigation in narrow hallways~\cite{park2023learning} and to augment existing global motion plans for which the global path is optimal~\cite{das2024motion}.

However, the learned hallucination approaches~\cite{wang2021agile, ghani2025dyna} exhibit a limitation that constrains their generalization. LfLH~\cite{wang2021agile} hallucinates obstacles by learning a distribution over the location and elliptical dimensions of static obstacles. 
Dyna-LfLH~\cite{ghani2025dyna} extends this model to dynamic environments by simplifying obstacle geometry and introducing a velocity component. 
Although both formulations are sound in principle, their learnable components are prone to mode collapse when paired with the fixed, optimal decoder. LfLH partially mitigates this effect in static environments by hallucinating a larger number of obstacles, increasing the diversity of the resulting configurations. 
In dynamic settings, however, densely populating the environment with hallucinated moving obstacles tends to enclose the robot within a collision-free corridor, so the training data no longer exposes the planner to genuine avoidance decisions.


\acronym\ avoids this failure mode by not learning a generative model over obstacle trajectories at all. Instead, it learns only \emph{where} and \emph{when} an obstacle must exist for the plan to be near-optimal, and generates trajectories through these critical points procedurally. Decoupling the learned component from trajectory synthesis removes the pressure toward mode collapse and yields obstacle diversity without having to densely populate the environment.

\begin{figure*}[t]
  \centering
  \begin{subfigure}{0.225\textwidth}
    \includegraphics[width=\linewidth,height=5.5cm,keepaspectratio]{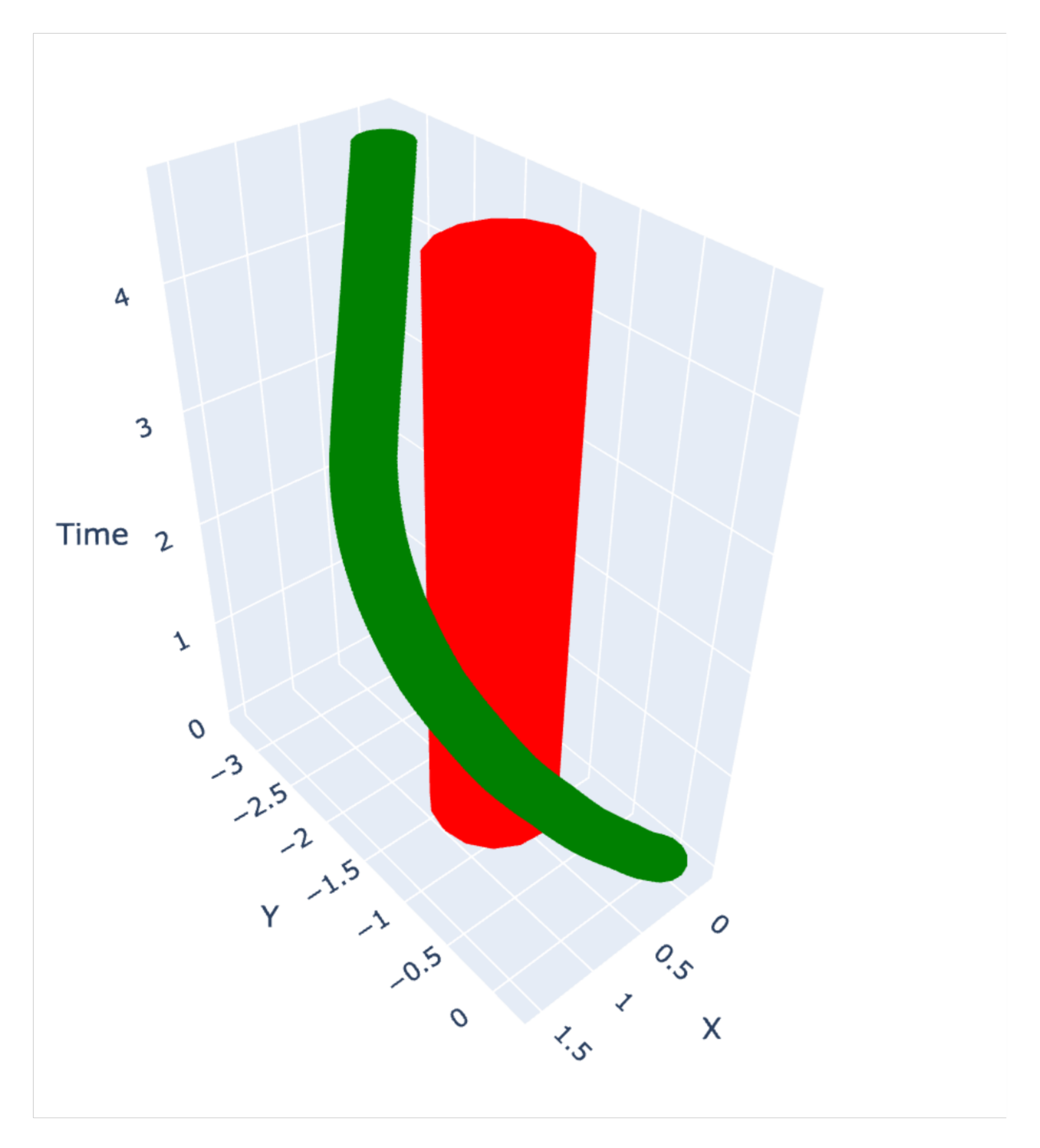}
    \caption{}
    \label{fig:a}
  \end{subfigure}
  \begin{subfigure}{0.225\textwidth}
    \includegraphics[width=\linewidth,height=5.5cm,keepaspectratio]{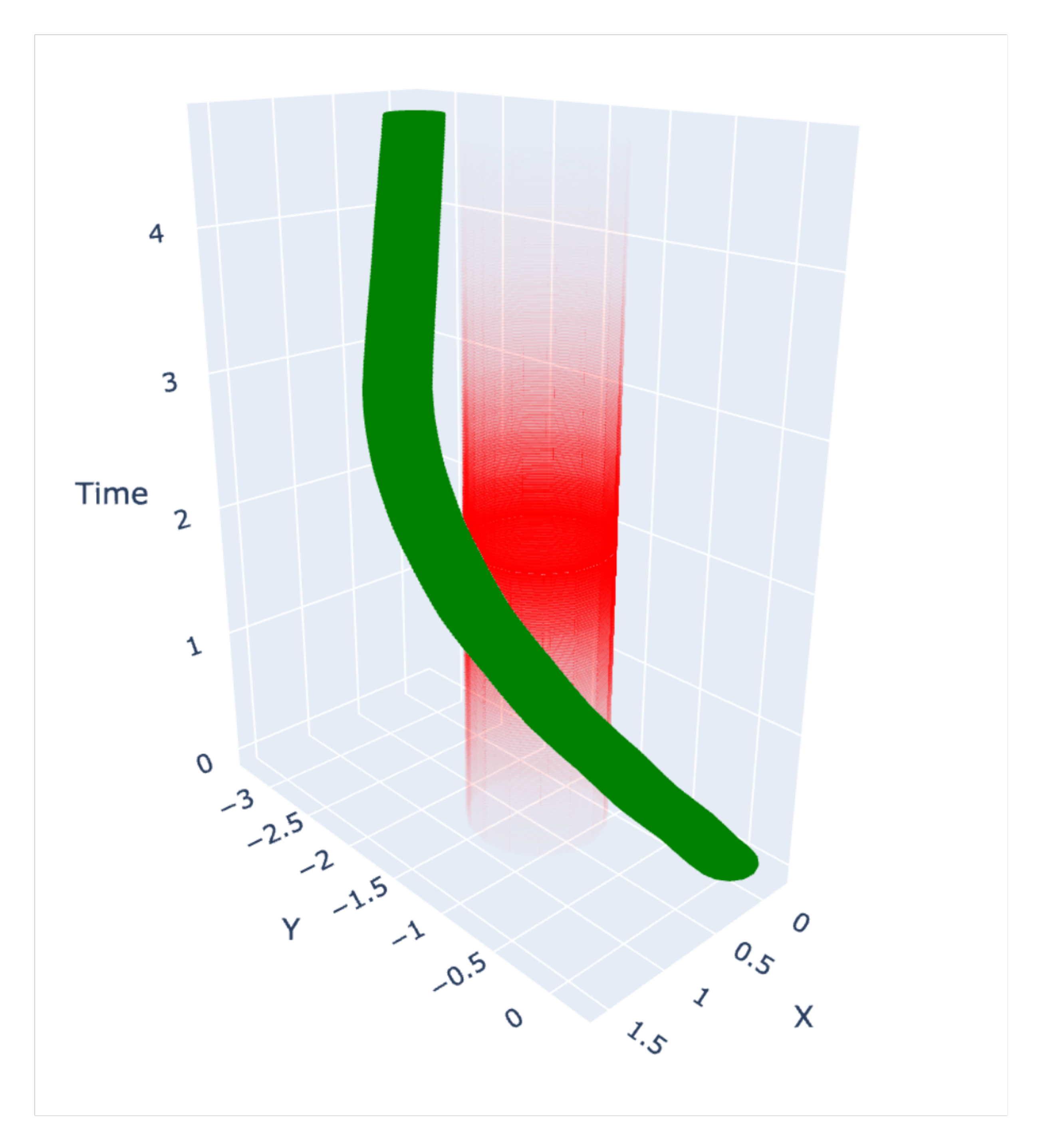}
    \caption{}
    \label{fig:b}
  \end{subfigure}
  \begin{subfigure}{0.225\textwidth}
    \includegraphics[width=\linewidth,height=5.5cm,keepaspectratio]{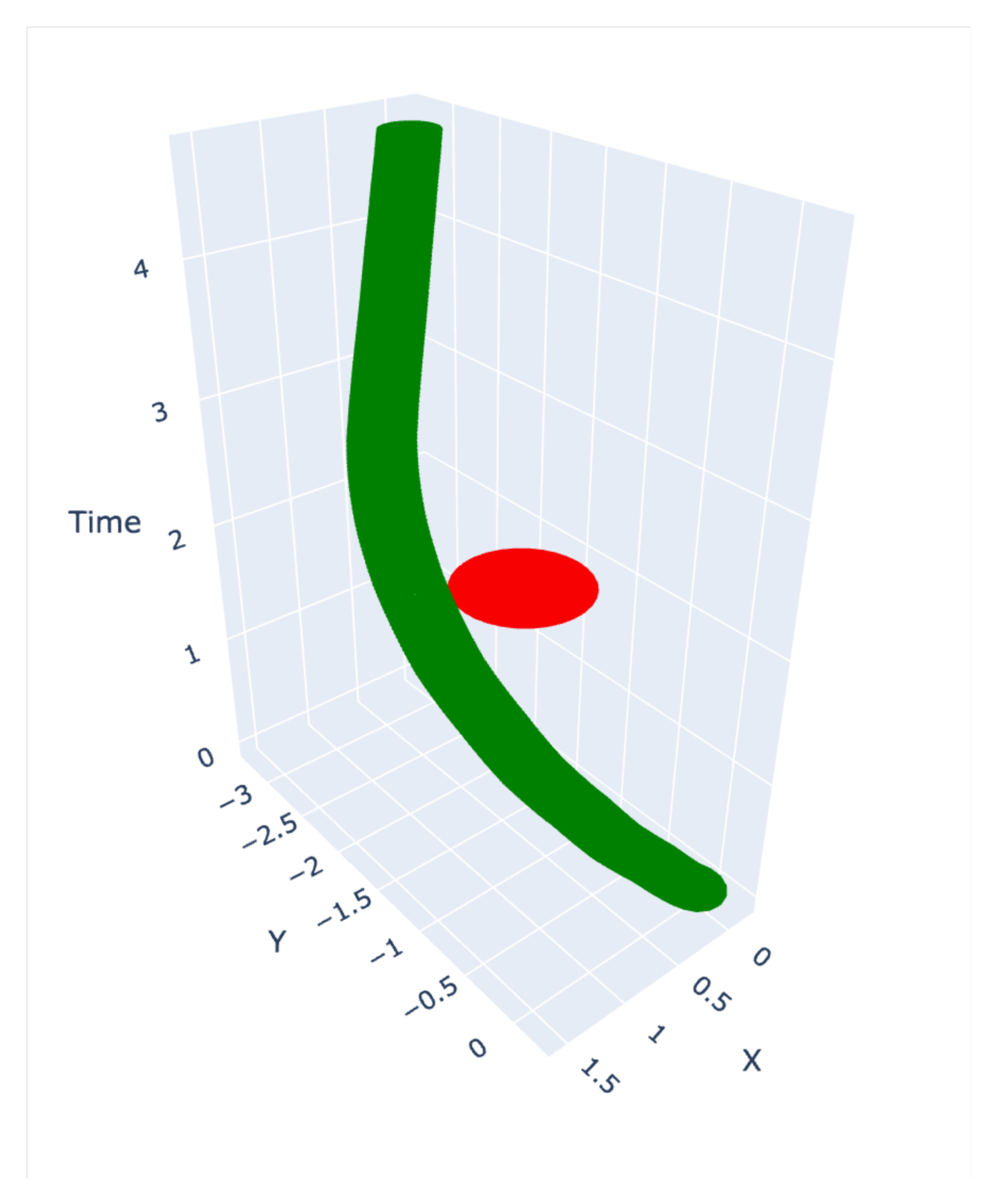}
    \caption{}
    \label{fig:c}
  \end{subfigure}
  \begin{subfigure}{0.3\textwidth}
    \includegraphics[width=\linewidth,height=5.5cm,keepaspectratio]{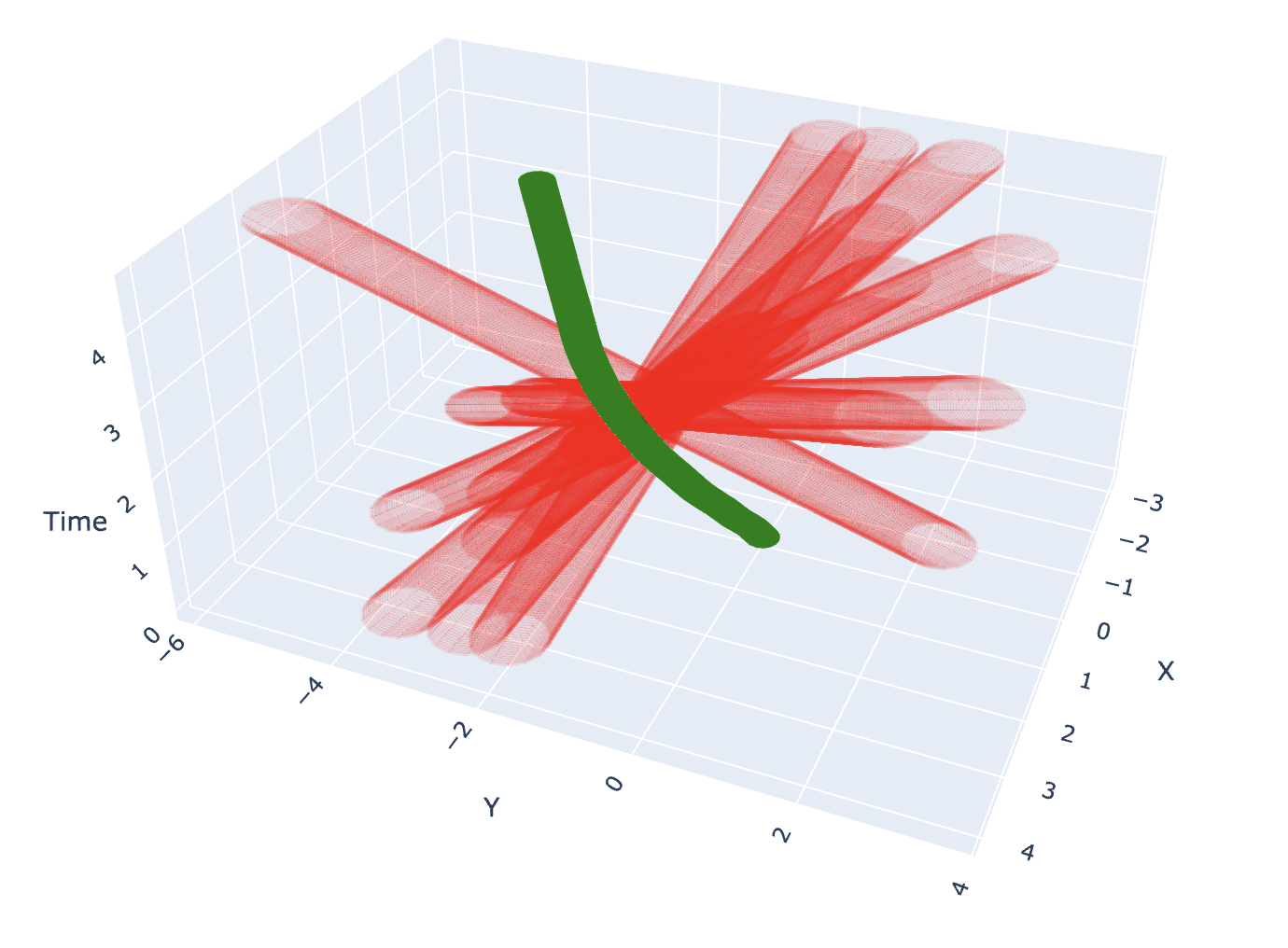}
    \caption{}
    \label{fig:d}
  \end{subfigure}
  \caption{
  \small
  The \acronym~method visualized for motion plan (green) and one obstacle (red). We first learn where the obstacle should be to make our plan optimal using Eqn.~\ref{eqn::psi_phase1} (Fig.~\ref{fig:a}). Then using Eqn.~\ref{eqn::psi_phase2}, we learn where the obstacle should be temporally to make the plan near-optimal (Fig.~\ref{fig:b},\ref{fig:c}). Finally using Eqn.~\ref{eqn::generation_function}, numerous obstacle trajectories can be generated from the obstacle's critical point (Fig.~\ref{fig:d}).}
  \label{fig::approach}
\end{figure*}

\section{Approach}
\label{sec::approach}

\begin{figure*}[t]
  \centering
  \includegraphics[width=2.0\columnwidth]{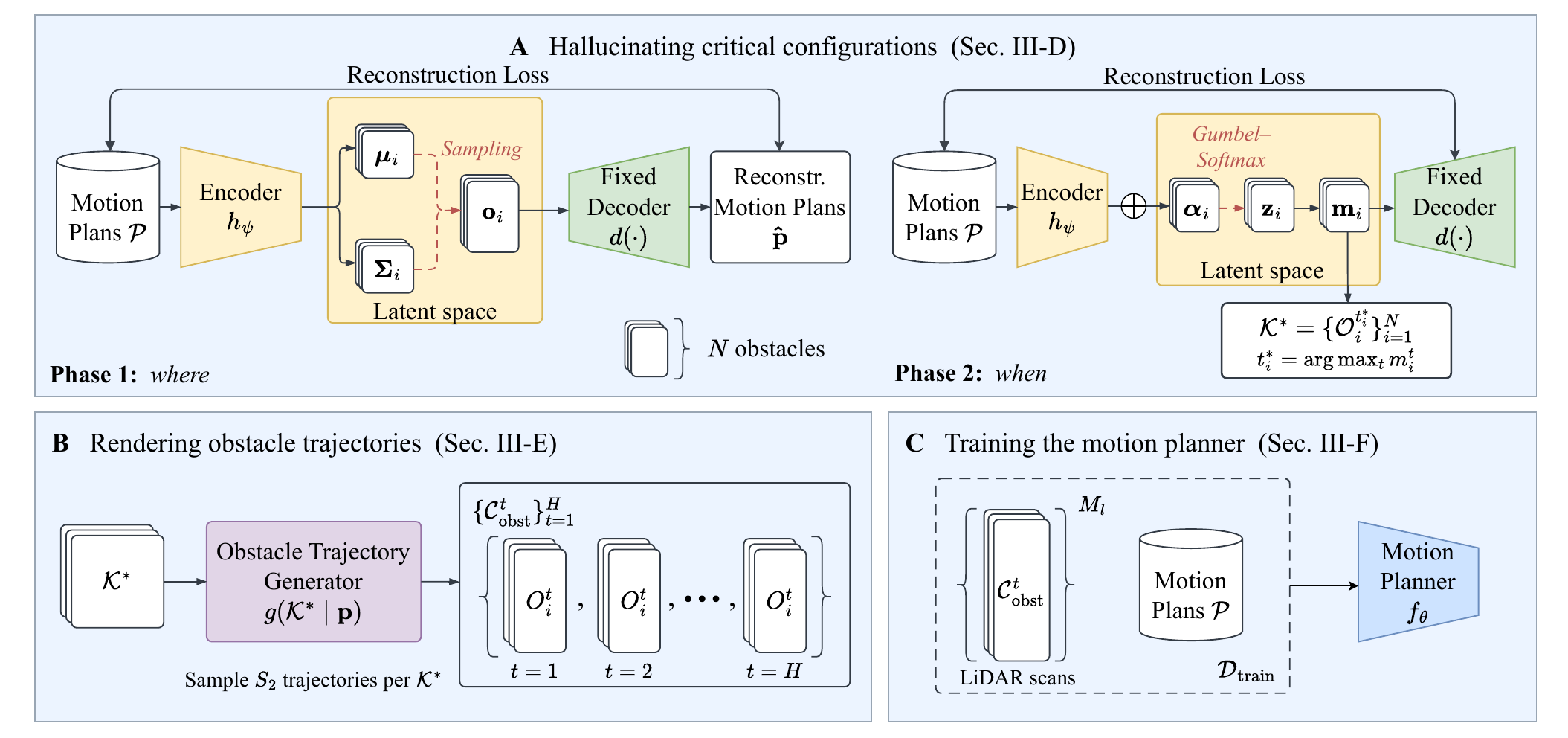}
  \caption{
\small
\textbf{A.} The encoder-decoder architecture learns the critical point configuration $\mathcal{K}^*$ over two phases. 
\textbf{Phase 1} (\emph{where}) learns a distribution over obstacle locations $\{\mathbf{o}_i\}^N_{i=1}$ that make $\mathbf{p} \in \mathcal{P}$ optimal. 
\textbf{Phase 2} (\emph{when}) additionally learns a temporal mask per obstacle via Gumbel--Softmax, whose peak gives the critical time step $t^*_i$. 
\textbf{B.} The generation function $g(\cdot)$ generates $S_2$ collision-free obstacle trajectories per critical point over horizon $H$. 
\textbf{C.} A motion planner $f_\theta(\cdot)$ is trained on $\mathcal{D}_\mathrm{train}$: $M_l$ past LiDAR scans paired with the motion plans $\mathbf{p}$.}
  \label{fig::architecture}
\end{figure*}

In this section, we begin by presenting the motion planning problem in dynamic environments and introduce the notion of \textit{critical configurations}. Using this notion, we reformulate the problem under the Learning from Hallucination (LfH) paradigm, which provides  the foundation for our proposed \acronym\ approach, visualized in Fig.~\ref{fig::approach} and \ref{fig::architecture}.

\subsection{The Motion Planning Problem}

Motion planning is often framed in the configuration space (C-space), which represents all possible robot configurations in a given environment. It is split into $\mathcal{C}_{\mathrm{obst}}$, the set of infeasible configurations blocked by obstacles or restricted by kinodynamic constraints, and $\mathcal{C}_{\mathrm{free}} = \text{C-space}  \setminus \mathcal{C}_{\mathrm{obst}}$, the set of feasible configurations. 

In dynamic environments, the C-space evolves over time, yielding $\mathcal{C}_{\mathrm{obst}}^t$ and $\mathcal{C}_{\mathrm{free}}^t$ where $t \in \{1, \dots, H\}$ over a discrete time horizon $H$. 
Only the obstacle-induced component of $\mathcal{C}_{\mathrm{obst}}^t$ varies with $t$; the kinodynamic constraints depend on the robot alone and are identical at every time step. For the remainder of the paper, $\mathcal{C}_{\mathrm{obst}}^t$ denotes the
obstacle-induced component alone.

A motion plan is a sequence of actions that moves the robot from its current configuration $\mathbf{c}_c$ to a goal configuration $\mathbf{c}_g$ through intermediate configurations $\mathbf{c}^t$ such that $\mathbf{c}^t \in \mathcal{C}_{\mathrm{free}}^t, \forall t$.  
The planning problem is then to find a function that generates such feasible plans. 
Formally, 
$$\mathbf{p} = f(\{\mathcal{C}_{\mathrm{obst}}^t\}^H_{t=1} \mid \mathbf{c}_c, \mathbf{c}_g),$$
where $\mathbf{p} = [\mathbf{u}^t]^{H-1}_{t=0}$,  $\mathbf{u}^t \in \mathcal{U}$, and $\mathcal{U}$ denotes the robot's action space. Kinodynamic feasibility is enforced through $\mathcal{U}$.
An optimal planner $f^*(\cdot)$ seeks an optimal plan $\mathbf{p}^*$ that minimizes a given cost, such as travel time or path length.

\subsection{The Critical Configurations}
We posit that an optimal plan $\mathbf{p}^*$ does not depend on the full time-varying obstacle configurations $\{\mathcal{C}_{\mathrm{obst}}^t\}^H_{t=1}$, but rather on a smaller subset of time steps, which we term the \textit{critical configurations}, denoted as $\mathcal{K}$:
\begin{equation}
\mathcal{K}=\{\mathcal{C}_{\mathrm{obst}}^t \colon t \in \mathcal{T}\}, \quad \mathcal{T} \subseteq \{1,\dots,H\},
\label{eqn::critical_configuration}
\end{equation}
where $\mathcal{T}$ represents the critical time step(s). At all other time step(s) the obstacle-induced constraints are dropped entirely.

The underlying intuition is that an obstacle constrains the plan only while it is near the robot. At all other times, it imposes no active constraint and can be omitted without affecting the plan's optimality. This motivates retaining the obstacle configurations only at the time steps where they actively constrain the plan, and treating the remainder of the horizon as configuration space free of any obstacle-induced constraints. Consequently, the plan considers obstacles only at time steps in $\mathcal{T}$.

Formally, the planning problem can be reformulated as 
$$\mathbf{p}^* = f^*(\mathcal{K} \mid \mathbf{c}_c, \mathbf{c}_g).$$
This reformulation, however, raises the non-trivial problem of identifying the critical configurations $\mathcal{K}$ from $\{\mathcal{C}_\mathrm{obst}^t\}^H_{t=1}$.
To address this, we leverage the LfH paradigm which enables direct synthesis of  the critical configurations $\mathcal{K}$ from past motion plans $\mathbf{p}$.

\subsection{Reformulation in LfH paradigm}

The LfH paradigm provides a natural way to tackle the challenge of identifying critical configurations by solving the inverse problem: given a motion plan, hallucinate the obstacle configurations for which
that plan is optimal. Concretely, a set of motion plans
$\mathcal{P}$ is collected from prior robot operation in obstacle-free space, and each $\mathbf{p} \in \mathcal{P}$ is treated as a demonstration to be explained.

Formally in dynamic environments this is expressed as:
$$\{\{\mathcal{C}_{\mathrm{obst}}^{t,k}\}_{t=1}^H\}_{k=1}^\infty = f^{-1}(\mathbf{p} \mid  \mathbf{c}_c, \mathbf{c}_g),$$ 
that is, hallucinating all (possibly infinite) obstacle configuration sequences over time horizon $H$ that make $\mathbf{p}$ optimal, given that $\mathbf{p}$ moves the robot from $\mathbf{c}_c$ to $\mathbf{c}_g$.

In our reformulation, LfH would focus on hallucinating the \textit{critical configurations} directly: 
\begin{equation}
    \{\mathcal{K}^k\}_{k=1}^\infty = f^{-1}(\mathbf{p} \mid  \mathbf{c}_c, \mathbf{c}_g),
    \label{eqn::raw_hallucination}
\end{equation}
representing all critical obstacle configurations that determine plan optimality.

Since it is impossible to predict all (infinite) possible configurations, we approximate it using a learned distribution. Concretely, the hallucination function \saad{$h(\cdot)$} outputs a distribution over the critical configurations:
\begin{equation}
    \mathcal{K} \sim h(\mathbf{p} \mid  \mathbf{c}_c, \mathbf{c}_g).
    \label{eqn::hallucination}
\end{equation}
This formulation forms the basis of our \acronym\ approach.

\subsection{Hallucinating Critical Configurations}

Following prior work, we approximate the obstacle-induced configuration space over the horizon by $N$ circular obstacles of fixed radius, writing $\mathcal{O}^t_i$ for the region occupied by obstacle $i$ at time $t$ and $\mathbf{o}_i^t$ for its
center.
The obstacle-induced configuration space then decomposes per obstacle in space-time,
$$
\{\mathcal{C}^t_\mathrm{obst}\}^H_{t=1}
\approx \{\mathcal{O}_i^t \colon i \in \{1,\dots,N\};~t \in \{1,\dots,H\}\},
$$
and the critical configurations are the restriction of this set to each obstacle's critical time steps,
\begin{equation}
\mathcal{K} \approx \big\{\, \mathcal{O}_i^t \colon
i \in \{1,\dots,N\};~ t \in \mathcal{T}_i \,\big\},
\quad \mathcal{T} = \bigcup\nolimits_{i=1}^{N} \mathcal{T}_i ,
\label{eqn::critical_decomposition}
\end{equation}
where $\mathcal{T}_i \subseteq \{1,\dots,H\}$ are the critical time steps of obstacle $i$.

The hallucination function $h_\psi(\cdot)$, parameterized by $\psi$, is learned in an encoder--decoder manner over two phases: the first determines
\textit{where} obstacles lie, the second \textit{when} they are present.

\textbf{Phase 1: where.}
Obstacles are held present at all time steps, so we drop the time index and
write $\mathcal{O}_i$ for the region occupied by obstacle $i$ throughout the
horizon. Each center is modeled as Gaussian,
$$
\mathbf{o}_i \sim \mathcal{N}(\bm{\mu}_i, \bm{\Sigma}_i), \qquad i = 1,\dots,N,
$$
and the encoder predicts these parameters from the motion plan and the current
and goal configurations,
\begin{equation}
\{(\bm{\mu}_i, \bm{\Sigma}_i)\}_{i=1}^{N}
= h_\psi(\mathbf{p} \mid \mathbf{c}_c, \mathbf{c}_g).
\label{eqn::encoder_phase1}
\end{equation}

To learn the distributions, the decoder $d(\cdot)$ is a classical motion planner with no learnable parameters that returns an optimal plan given a sampled obstacle set,
$\mathcal{O} \triangleq \{\mathcal{O}_i\}_{i=1}^{N}$. We fit $\psi$ by minimizing the reconstruction loss,
\begin{equation}
\psi^* = \argmin_\psi \mathop{\mathbb{E}}_{\substack{\mathbf{p} \sim \mathcal{P}\\\mathcal{O} \sim h_\psi(\mathbf{p} \mid \mathbf{c}_c, \mathbf{c}_g) }} \ell(\mathbf{p}, d(\mathcal{O} \mid \mathbf{c}_c, \mathbf{c}_g)),
\label{eqn::psi_phase1}
\end{equation}
so that the hallucinated obstacles make $\mathbf{p}$ optimal.

\textbf{Phase 2: when.}
We now estimate $\mathcal{T}_i$ for each obstacle. Alongside the outputs of
Eqn.~\eqref{eqn::encoder_phase1}, $h_\psi(\cdot)$ emits logits
$\bm{\alpha}_i \in \mathbb{R}^H$ per obstacle. Passing logits $\bm{\alpha}_i$ through a Gumbel--Softmax yields $\mathbf{z}_i \in [0,1]^H$ with $\sum_{t} z_i^t = 1$, a
relaxed one-hot vector over the $H$ time steps. Normalizing by its largest
entry gives a soft, continuous mask
$$
\mathbf{m}_i = \frac{\mathbf{z}_i}{ \max\limits_{t} z_i^t} \;\in [0,1]^H,
$$
whose entries encode the temporal presence of obstacle $i$: $m_i^t = 0$ denotes
absence and $m_i^t = 1$ full presence at time step $t$.

In this work, we restrict $|\mathcal{T}_i| =1$, retaining only the single time step at which the obstacle most strongly constrains the plan.
This extreme temporal restriction relaxes the problem, so that $\mathbf{p}$ is recovered as near-optimal rather than optimal, in exchange for the temporal sparsity that makes the trajectory generation of Sec.~III-E possible. We denote the pair
$(\mathbf{o}_i, t_i^*)$ as the \textit{critical point} of obstacle $i$, formed by its center and the time step at which its mask is maximal,
\begin{equation}
\begin{aligned}
\{(\bm{\mu}_i, \bm{\Sigma}_i, \bm{\alpha}_i)\}_{i=1}^{N}
&= h_\psi(\mathbf{p} \mid \mathbf{c}_c, \mathbf{c}_g), \\
t_i^* &= \argmax_{t} m_i^t,
\end{aligned}
\label{eqn::critical_point}
\end{equation}
so that each obstacle exists at exactly one time step. This reduces Eqn.~\eqref{eqn::critical_decomposition} to the $N$ obstacle regions
evaluated only at their critical points, resulting in the critical point configuration and a singleton critical time step of obstacle $i$, 
\begin{equation}  
\mathcal{K}^* = \big\{\, \mathcal{O}_i^t \colon
i \in \{1,\dots,N\};~ t  = t^*_i \big\}.
\label{eqn::critical_point_configuration}
\end{equation}

For training, the hard assignment in Eqn.~\eqref{eqn::critical_point_configuration} is
replaced by the soft, differentiable set
$\mathcal{K}_{\mathrm{soft}} \triangleq
\{(\mathcal{O}_i, \mathbf{m}_i)\}_{i=1}^{N}$, which retains the full mask and
lets the decoder account for partial obstacle presence (see Sec. \ref{subsec::impl_h_fn}). The objective is
otherwise unchanged:
\begin{equation}
    \psi^* = \argmin_\psi \mathop{\mathbb{E}}_{\substack{\mathbf{p} \sim \mathcal{P}\\\mathcal{K}_\mathrm{soft}\sim h_\psi(\mathbf{p} \mid \mathbf{c}_c, \mathbf{c}_g) }} \ell(\mathbf{p}, d(\mathcal{K}_\mathrm{soft} \mid \mathbf{c}_c, \mathbf{c}_g)).
    \label{eqn::psi_phase2}
\end{equation}

\subsection{Rendering Obstacle Trajectories}
The generation function $g(\mathcal{K}^{*}\!\mid\!\mathbf{p})$ uses the critical point configuration to generate arbitrarily many obstacle trajectories over the horizon $H$. Specifically,
\begin{equation}
\begin{split}
     \{\mathcal{C}_\mathrm{obst}^t\}^H_{t=1} &\approx g(\mathcal{K}^*\mid \mathbf{p}), \\
     g(\mathcal{K}^*\mid \mathbf{p}) &=  \big\{\mathcal{O}_i^t \colon
i \in \{1,\dots,N\};~ t \in \{1,\dots,H\}\big\},
     \label{eqn::generation_function}
\end{split}
\end{equation}
that is, a full space-time trajectory for every obstacle.

$g(\mathcal{K}^*\mid \mathbf{p})$ is not tied to a specific parameterization, but it must satisfy two constraints: (i) each obstacle center must be at its critical location at its critical time, $\mathbf{o}_i^{t} = \mathbf{o}_i,\hspace{0.5em} t=t^*_i$; and (ii) the generated trajectories must remain collision-free with the plan $\mathbf{p}$. Examples of $g(\cdot)$ are functions that sample smooth trajectories from first order dynamics (see Eqn. \ref{eqn::instantiate_gen_fn}), or piece-wise functions that define obstacle motion over the horizon. Fig.~\ref{fig::generated_obstacle_trajectories} visualizes obstacle trajectories generated from $N=2$ and $N=3$ critical points.

\subsection{Learning from Hallucinating Critical Points}

For every plan $\mathbf{p} \in \mathcal{P}$ we sample $S_1$ sets of critical points from $h_{\psi^*}(\cdot)$ and, from each, generate $S_2$ obstacle
trajectories with $g(\cdot)$, yielding $S = |\mathcal{P}| \, S_1 S_2$
supervised training instances for IL,
$$
\mathcal{D}_{\mathrm{train}} = \big\{(\{\mathcal{C}^{t,j}_{\mathrm{obst}}\}^H_{t=1}, \mathbf{p}^j, \mathbf{c}_c^j, \mathbf{c}_g^j ) \colon~ j \in \{1,\dots,S\}\big\}.
$$
Here, each plan $\mathbf{p}^j$ is (roughly) optimal for its corresponding configuration space $\{\mathcal{C}^{t,j}_\mathrm{obst}\}^H_{t=1}$. The configuration space is transformed into sensor observations (e.g., LiDAR scans via ray tracing), enabling us to train a motion planner $f_\theta(\cdot)$ to reproduce $\mathbf{p}^j$ from $\{\mathcal{C}^{t,j}_{\mathrm{obst}}\}_{t=1}^H$. Formally,
\begin{equation}
\small
\theta^* = \argmin_\theta \mathop{\mathbb{E}}_{\substack{(\{\mathcal{C}^t_\mathrm{obst}\}^H_{t=1}, \mathbf{p}, \mathbf{c}_c, \mathbf{c}_g)\\ \sim \mathcal{D}_\mathrm{train}}} \big[ \ell(\mathbf{p}, f_\theta(\{\mathcal{C}^t_\mathrm{obst}\}^H_{t=1} \mid \mathbf{c}_c, \mathbf{c}_g))\big].
\label{eqn::theta}
\end{equation}
During deployment, $f_{\theta^*}(\cdot)$ will be used to plan around the sensed dynamic environment.

\section{Implementation}

In this section we discuss the particular instantiation of all data and models discussed in Sec. \ref{sec::approach}.

\subsection{Learning $h_\psi$}
\label{subsec::impl_h_fn}
We collect a dataset of motion plans $\mathcal{P}$, where each plan is a sequence of $\mathrm{SE}(2)$ robot configurations $(x^t,y^t,\phi^t)$ and linear and angular velocity actions $(v^t,\omega^t)$. Each plan is segmented using a sliding window of size 233 ($\approx$5\,s of odometry time). Configurations are expressed in the robot frame, so $\mathbf{c}_c=\mathbf{0}$ and is omitted. $\mathbf{c}_g$ is set as the final configuration of each segment and is not modeled separately.

The hallucination function $h_\psi$ is implemented as a 3-layer convolutional encoder (channels [16,32,64], kernels [5,5,3], stride 2), followed by $N$ autoregressive linear layers that output $\bm{\mu}$, $\bm{\Sigma}$, and $\bm{\alpha}$.

Training proceeds in two phases. In Phase 1 of training, $h_\psi$ is optimized for 1000 epochs to predict $\bm{\mu}$ and $\bm{\Sigma}$, generating $N=10$ circular obstacles (radius 0.5\,m) centered at sampled $\{(x_i,y_i)\}_{i=1}^N$ coordinates. In Phase~2, the network learns $\bm\alpha$ over 1500 epochs. Initially, obstacle temporal masks $\mathbf{m}_i = \mathbf{1}$, i.e. obstacle is present at all time steps. Over 1000 epochs, $\mathbf{m}_i$ is annealed to one-hot vectors at $t_i^*=\argmax \mathbf{m}_i$ via a Gumbel–Softmax with temperature $\tau$ decaying from 2048 to 0.1. The final 500 epochs train with one-hot $\mathbf{m}_i$.

The decoder $d$ is a re-implementation of Ego-Planner~\cite{zhou2020ego} with convex optimization layers. It uses $\mathbf{m}_i$ to scale obstacle radii and safety clearances over time, so that $m_i^t \to 0$ nullifies collisions with obstacle $i$ at time $t$. 

The reconstruction loss is mean-squared error between the input plan $\mathbf{p}$ and reconstructed plan $\mathbf{p}^*$. Following prior methods, we stabilize training by adding priors and penalties: (i) a Gaussian prior fitted to $\mathbf{p}$ is imposed on obstacle centers, biasing obstacles toward the trajectory; and (ii) penalties are applied for obstacle–obstacle and obstacle–plan overlap.


\subsection{Generating $\mathcal{D}_\mathrm{train}$}

For generating $\mathcal{D}_\mathrm{train}$, we sample $S_1=1$ critical point per obstacle from $h_{\psi^*}$ for each $\mathbf{p}^j \in \mathcal{P}$, forming $\mathcal{K}^{*j}$. To isolate only the obstacles essential for optimizing the plan, we filter based on reconstruction loss: obstacles are added incrementally in order of contribution, and retained only if they reduce the loss by at least $1\%$. The process stops when no further improvement is observed or when $N_{\mathrm{max}}=7$ is reached.
Finally, we require at least a $90\%$ overall reduction in reconstruction loss relative to the baseline straight-line plan from start to goal, obtained when obstacles are unoptimized and placed far away.
The resulting $\mathcal{K}^{*j}_{\mathrm{filtered}}$ thus contains only the obstacle critical points that contribute most to optimizing $\mathbf{p}^j$.

Obstacle trajectories are instantiated using $g(\mathcal{K}^{*j}_\mathrm{filtered}\mid \mathbf{p}^j)$, which samples velocities from a uniform distributions between $[1, 2]$ m/s, and applies the first-order equation of motion to produce smooth trajectories. Collision checks ensure that the generated trajectories do not intersect the corresponding plan $\mathbf{p}^j$:
{\small
\begin{equation}
\begin{aligned}
g\Big(\{\mathcal{O}_i^{t^*_i}\}_{i=1}^{|\mathcal{K}^{*j}_\mathrm{filtered}|} \mid \mathbf{p}^j\Big) &= 
\Big\{\mathbf{o}_i + (t-t^*_i)\mathbf{v}_i \Big\},\\
&t \in \{1,\dots, H\},\\ & \forall i = \{1, \dots, |\mathcal{K}^{*j}_\mathrm{filtered}|\},
\label{eqn::instantiate_gen_fn}
\end{aligned}
\end{equation}}
where $\mathbf{v}_i$ is the $(x,y)$ components of the sampled velocity.

Using Eqn.~\eqref{eqn::instantiate_gen_fn}, we sample $S_2=50$ trajectories from each $\mathcal{K}^{*j}_\mathrm{filtered}$, producing a total of $|\mathcal{P}|\times50$ training instances, each containing up to 7 obstacles. These are rendered as 2D LiDAR scans given $\mathbf{c}^{t,j}$ along $\mathbf{p}^j$.

Similar to prior methods~\cite{ghani2025dyna}, $\mathcal{D}_\mathrm{train}$ is further augmented to improve robustness. First, up to 20 random non-colliding obstacle trajectories are added to $\mathcal{C}_\mathrm{obst}$ to introduce noisy, non-optimizing obstacles. Second, obstacle-free motion plans with speed above $0.9\,\text{m/s}$ and direction aligned with the goal $90\%$ of the time (cosine similarity $\geq 0.9$) are added to encourage fast goal-directed navigation in open spaces.

\subsection{Learning and Deployment of $f_{\theta^*}$}

The motion planner $f_{\theta^*}$ predicts a sequence of $M_a=5$ actions, $\{\mathbf{u}^t\}_{t=1}^{M_a}$, conditioned on $M_l=5$ LiDAR scans (comprising $M_l-1$ historical scans and the current scan) representing obstacle configurations $\{\mathcal{C}^t_{\mathrm{obst}}\}_{t=t-M_l+1}^{t}$ and on $M_l=5$ past actions. 
Using multiple past scans ($M_l>1$) enables the planner to capture obstacle dynamics, while predicting multiple future actions ($M_a>1$) supports longer-horizon reasoning. 
Each training instance is anchored at the robot's current configuration $\mathbf{c}_c^t=\mathbf{0}$, with the goal $\mathbf{c}_g^t$ defined as a unit vector pointing to the end of plan $\mathbf{p}$.

The planner $f_{\theta^*}(\cdot)$ is implemented as a causal transformer with two encoder layers, two attention heads, and a 256-dimensional feed-forward network. LiDAR scans are first projected into a 256-dimensional latent space using a 2-layer 1D CNN (kernel size 5). The transformer output is concatenated with the goal vector and processed by a 2-layer MLP head to predict linear and angular velocities.

During deployment, at each time step $t$, the robot’s current configuration is set as $\mathbf{c}_c^{t}=\mathbf{0}$, and the goal $\mathbf{c}_g^{t}$ is a unit vector pointing $2.25\,\text{m}$ ahead along the global path given by the \texttt{move\_base} navigation stack. 
The planner predicts $M_a$ actions but executes only the first at each step. Following prior work, an MPC safety module monitors collisions: if a collision is imminent, the robot halts first and then reverses to avoid obstacles.

\section{Experiments and Evaluation}


\label{sec::experiments}
In this section, we first demonstrate that a single critical point is sufficient to produce a near-optimal trajectory. We then evaluate the hallucination of critical points and introduce a metric for dataset richness, showing how our method generates diverse obstacle trajectories that maximize this metric. Finally, we present results for learning a motion planner from the hallucinated trajectories and demonstrate that the learned planner outperforms \saad{a baseline method} in simulated environments.

\subsection{Hallucinating Critical Points}

\begin{figure}
  \centering
  \includegraphics[width=1.0\columnwidth]{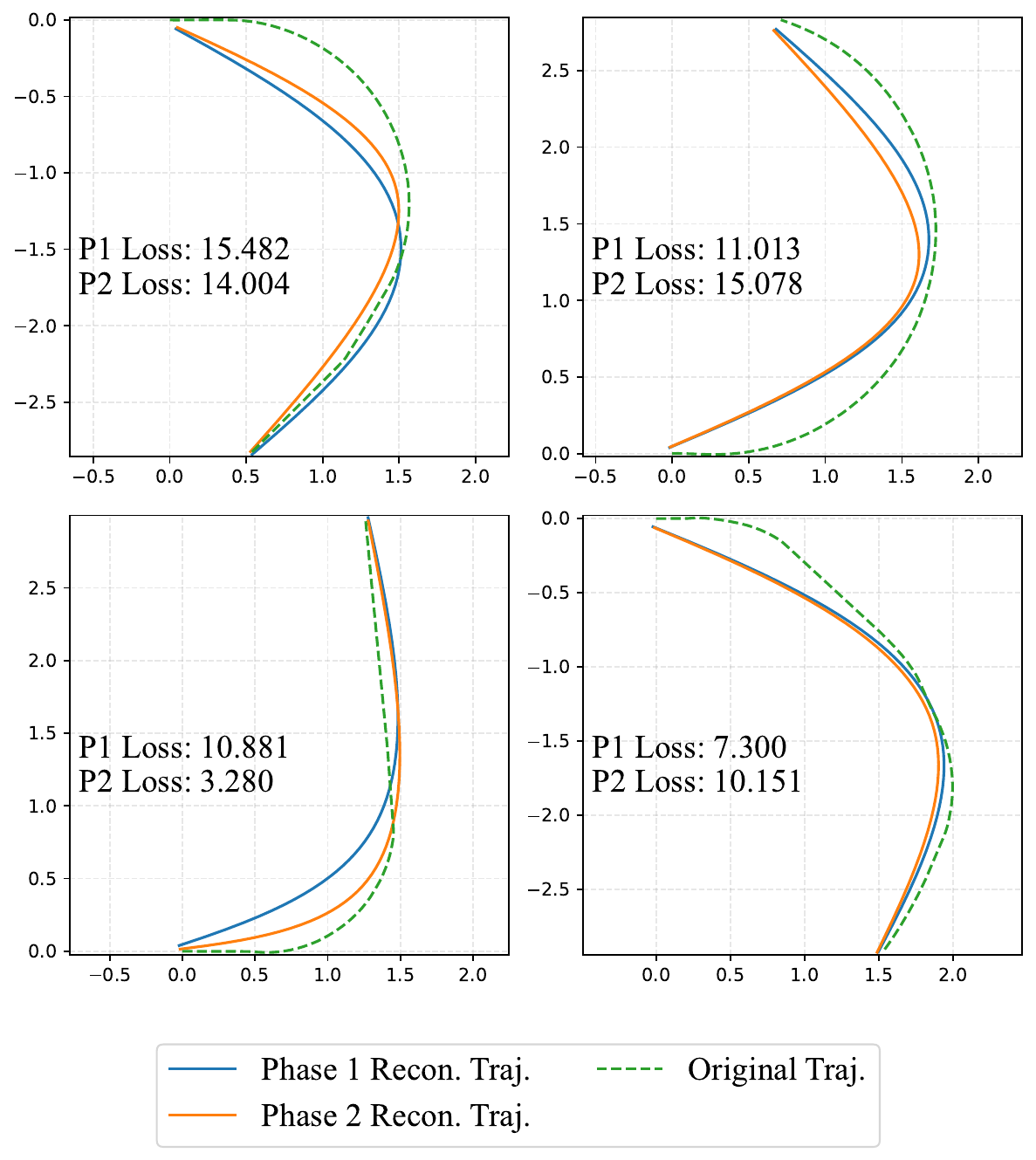}
  \caption{Reconstructed trajectories from Phase~1 (P1) and Phase~2 (P2) using obstacles at their critical points. Trajectories remain visually close to each other, with only minor variations in reconstruction loss.}
  \label{fig::recon_trajectories}
\end{figure}

In Fig.~\ref{fig::recon_trajectories}, we compare the original trajectories with those reconstructed from obstacles at the end of Phase~1 (P1) and from obstacles instantiated solely at their critical points at the end of Phase~2 (P2). Visually, the P2 reconstructed trajectories remain similar to their P1 counterpart. Such similarities validate our hypothesis that obstacles only appearing at their critical points are sufficient to make a motion plan near-optimal, instead of requiring the obstacles' constant presence.

In Fig.~\ref{fig::generated_obstacle_trajectories}, we illustrate the diversity of obstacle trajectories that can be generated from a \textit{single} critical point. Despite the variations in their paths, all these obstacles can be avoided using the same underlying motion plan, which remains near-optimal in terms of travel efficiency.

\subsection{Measuring Dataset Coverage}
To quantify the richness of the generated datasets, we define a \textit{Dataset Coverage Score (DCS)}, which measures how well the dataset spans the space of relevant obstacle configurations. We consider four key components:
\begin{itemize}
\item Distance between the robot and obstacle centers, $r$;
\item Angle between the robot and obstacle centers, $\beta$; 
\item Obstacle speed, $v_\mathrm{o}$; and 
\item Obstacle heading in the robot frame, $\phi_\mathrm{o}$.
\end{itemize}

Each data point $\mathbf{z} = (r, \beta, v_\mathrm{o}, \phi_\mathrm{o})$ describes one
obstacle at one time step, counting only obstacles generated from critical points and
excluding the random obstacles added for robustness. Each component $c$ is discretized into
$B_c$ bins over the bounds and resolution given in
Tab.~\ref{tab::dataset_diversity_parameters}. For a subset of components $\mathcal{S}$,
coverage is measured over the product of their bins, indexed by $\mathbf{b}$: a bin is
occupied if at least one data point falls within it, and DCS is the fraction of bins
occupied,
\begin{equation}
\mathrm{DCS}(\mathcal{S}) = \frac{1}{\prod_{c \in \mathcal{S}} B_c}
\sum_{\mathbf{b}} \mathbf{1}[\,\mathbf{b} \text{ occupied}\,].
\label{eqn::dcs}
\end{equation}
A single component gives its marginal coverage; all four give the joint coverage over the
full space, which we report as the overall dataset coverage.

\begin{figure}
  \centering
  \includegraphics[width=0.8\columnwidth]{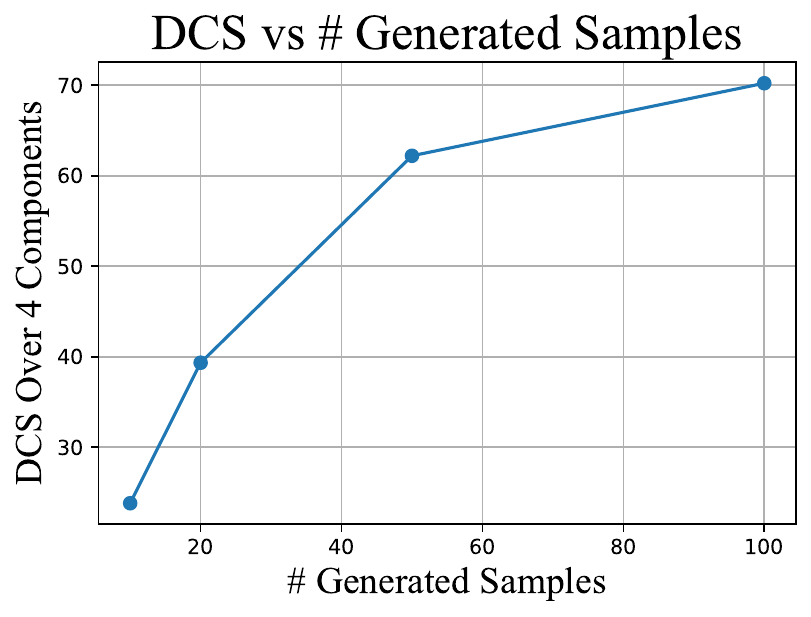}
  \caption{DCS increases as the number of samples per critical point, $S_2$, increases.}
  \label{fig::samples_vs_coverage}
\end{figure}

The coverage achieved by our method is reported in Tab.~\ref{tab::dataset_coverage}. \acronym~can achieve almost 100\% coverage when considering up to three metrics, while for all four metrics a coverage of 62.21\% is reached. In contrast, Dyna-LfLH achieves only 11.05\% over the same range. To further demonstrate robustness, Fig.~\ref{fig::samples_vs_coverage} shows how increasing the number of generated samples improves the coverage score. This indicates that our approach avoids mode collapse and effectively populates diverse obstacle scenarios.

\begin{table}[t]
\centering
\caption{Bounds and Resolution for Dataset Coverage.}
\begin{tabular}{cccc}
\toprule
Component (unit) & Lower bound & Upper bound & Resolution\\ 
\midrule
$r$ (m) & 0.15 & 2 & 0.1\\
$\beta$ (deg) & -180 & 180 & 5\\
$v_\mathrm{o}$ (m/s) & 1 & 2 & 0.1\\
$\phi_\mathrm{o}$ (deg) & -180 & 180 & 5\\
\bottomrule
\end{tabular}
\label{tab::dataset_diversity_parameters}
\end{table}

\begin{table}[t]
\centering
\caption{DCS for all component combinations for our \acronym\ and comparative approach.}
\begin{tabular}{ccc}
\toprule
 Component $\mathcal{S}$ & Coverage \acronym (\%) & Coverage Dyna-LfLH (\%) \\
\midrule
$r, \beta, v_\mathrm{o}, \phi_\mathrm{o}$ & 62.21 & 11.05\\
\midrule
$r, \beta, v_\mathrm{o}$ & 100.0 & 74.61 \\
$\beta, v_\mathrm{o}, \phi_\mathrm{o}$ & 99.18 & 49.30\\
$r, v_\mathrm{o}, \phi_\mathrm{o}$ & 100.0 &  69.6\\
$r, \beta, \phi_\mathrm{o}$ & 93.20 & 40.49\\
\midrule
$\beta, v_\mathrm{o}$ & 100.0 & 99.13 \\
$r, v_\mathrm{o}$ & 100.0 & 89.47 \\
$r, \beta$ & 100.0 & 89.40 \\
$r, \phi_\mathrm{o}$ & 100.0 & 86.52 \\
$\beta, \phi_\mathrm{o}$ & 100.0 & 89.06 \\
$v_\mathrm{o}, \phi_\mathrm{o}$ & 100.0 & 91.41 \\
\midrule
$r$ & 100.0 & 89.47 \\
$\beta$ & 100.0 & 100.0\\
$v_\mathrm{o}$ & 100.0 & 100.0\\
$\phi_\mathrm{o}$ & 100.0 & 100.0\\
\bottomrule
\end{tabular}
\label{tab::dataset_coverage}
\end{table}

\subsection{Benchmarking Navigation Performance}

We evaluate the learned motion planner using DynaBARN~\cite{nair2022dynabarn}, a simulation testbed for dynamic obstacle avoidance. DynaBARN supports generation of environments populated with moving obstacles at varying speeds, following trajectories with different motion patterns.

\begin{figure}
  \centering
  \includegraphics[width=1.0\columnwidth]{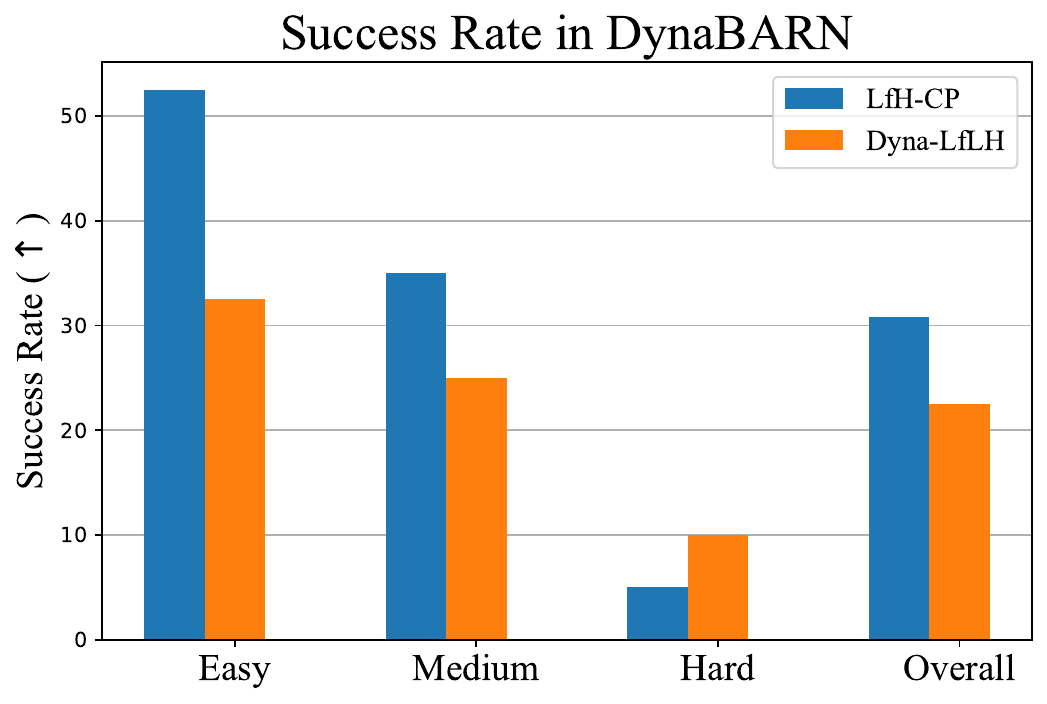}
  \caption{Simulation Results of 120 trials in DynaBARN.}
  \vspace{15pt}
  \label{fig::simulation_results}
  \vspace{-10pt}
\end{figure}

We generate 60 environments with 3 levels of difficulty similar to the original DynaBARN implementation using the parameters in Tab.~\ref{tab::dynabarn_parameters}. We compare \acronym~ against Dyna-LfLH. Each planner is evaluated over 2 trials in each of the 60 environments, yielding a total of 120 trials. Simulation results are summarized in Fig.~\ref{fig::simulation_results}.

\saad{Success rates in DynaBARN are low for both planners, reflecting the fast-moving and
highly cluttered environments. \acronym~succeeds in 30.83\% of trials against 22.5\% for
Dyna-LfLH, an absolute improvement of 8.33 points, and improves on the Easy and Medium
difficulties. On the Hard difficulty, however, \acronym~succeeds in fewer trials than
Dyna-LfLH. The overall gap is consistent with the difference in dataset coverage reported in
Tab.~\ref{tab::dataset_coverage}.}

\subsection{Limitations}
Despite achieving high dataset coverage, \acronym~succeeds in fewer than a third of DynaBARN trials. We identify three limitations. 
First, we restrict $\mathcal{T}$ to a single time step per obstacle.
Eqn.~\ref{eqn::critical_configuration} places no such restriction, and retaining a short temporal window would constrain the plan more tightly at the cost of reduced trajectory diversity. We adopt the single-point formulation to maximize diversity, but the trade-off between the two remains unexplored.

Second, each plan is optimized for only a few obstacles that are rendered at a time. While random non-colliding obstacles are added for robustness, the planner may still struggle when multiple obstacles appear simultaneously during deployment. Generating multiple obstacle samples passing through the same critical point within the same demonstration would produce denser training scenes and is a direct extension of the
present formulation. 

Third, DCS measures coverage for individual obstacles, requiring at least one sample per bin. During deployment however, multiple obstacles often appear together, suggesting that DCS could be extended to account for joint coverage of two or more obstacles simultaneously.

\begin{table}[t]
\centering
\caption{DynaBARN Parameters.}
\begin{tabular}{cccc}
\toprule
 Difficulty & \# Worlds  & Number of obstacles  & Range of speed \\ 
\midrule
Easy & 20 & [5, 10] & [0.5, 1.0] \\
Medium & 10 & [5, 10] & [1.0, 2.0] \\
Medium & 10 &[10, 20] & [0.5, 1.0] \\
Hard & 20 & [10, 20] & [1.0, 2.0] \\
\bottomrule
\end{tabular}
\label{tab::dynabarn_parameters}
\end{table}

\section{Conclusions}
\label{sec::conclusions}

In this paper, we present \acronym, a self-supervised framework for creating rich dynamic obstacle datasets based on existing optimal motion plans to create supervised training data. Situated within the LfH paradigm, \acronym\ does not require expensive expert demonstrations for IL or trial-and-error exploration for RL and extends to dynamic obstacles. Facing the vast and high-dimensional space of dynamic obstacles, \acronym\ further tackles the mode collapse problem that degraded the performance of the previous LfH approach by hallucinating only the critical points for those obstacles, i.e., where and when these obstacles have to appear to assure an existing motion plan's near-optimality. Our new obstacle diversity metric, DCS, shows that \acronym\ produces substantially more varied training data than existing baseline and DCS scales with more samples, instead of saturating due to mode collapse. \acronym\ also achieves improved navigation performance in dynamic obstacles compared to state-of-the-art LfH approach~\cite{ghani2025dyna}. 

\bibliographystyle{IEEEtran}
\bibliography{IEEEabrv,references}

\end{document}